\documentclass[letterpaper]{article}
\usepackage{arxiv}
\usepackage{times}
\usepackage{helvet}
\usepackage{courier}
\usepackage{url}
\usepackage{graphicx}
\usepackage{xcolor}

\title{Win Prediction in Esports: Mixed-Rank Match Prediction in Multi-player Online Battle Arena Games\thanks{This work is supported
by the Digital Creativity Labs jointly funded by EPSRC/AHRC/InnovateUK grant EP/M023265/1.}}
\author{Victoria Hodge, Sam Devlin, Nick Sephton, Florian  Block, Anders Drachen \and Peter Cowling\\
Digital Creativity Labs, University of York, UK\\
\{victoria.hodge, sam.devlin, nick.sephton, florian.block, anders.drachen, peter.cowling\}@york.ac.uk
}
\begin{document}
\maketitle
\begin{abstract}

Esports has emerged as a popular genre for players as well as spectators, supporting a global entertainment industry. Esports analytics has evolved to address the requirement for data-driven feedback, and is focused on cyber-athlete evaluation, strategy and prediction. Towards the latter, previous work has used match data from a variety of player ranks from hobbyist to professional players. However, professional players have been shown to behave differently than lower ranked players. Given the comparatively limited supply of professional data, a key question is thus whether mixed-rank match datasets can be used to create data-driven models which predict winners in professional matches and provide a simple in-game statistic for viewers and broadcasters. Here we show that, although there is a slightly reduced  accuracy, mixed-rank datasets can be used to predict the outcome of professional matches, with suitably optimized configurations.

\end{abstract}

\section{Introduction}\label{sec:Intro}
In the past decade, electronic sports (esports), has emerged as a popular genre for players and spectators and as a developing field for research ~\cite{schubert2016esports,superdata:17}. While there is no official definition of esports, the term fundamentally refers to digital games played in a competitive context with an audience. \citeauthor{superdata:17} predict that the esports market will be worth \$1.1 billion in 2017 and that there will be 330 million spectators by 2019 making esports an important research field across academia and industry.

The combination of esports' growth, and the availability of detailed data from virtually every match played, has given rise to the field of esports analytics ~\cite{schubert2016esports}. Many esports are complex and fast paced. Tactics and balance change much faster than in traditional sports. Providing simple, overarching statistics can help broaden the appeal of the games and make them more accessible to  viewers. \textbf{Win prediction} has formed the focal point of such analytics; it is a simple statistic easily understood by audience and players alike. Hence, a range of win prediction techniques have been developed, \cite{Semenov2017,2017arXiv170103162Y}. 

The focus of this paper is using the current game state to predict the likely winner for professional (tournament-level) games. Pro games have the highest industry and audience interest but are limited in number. We analyze the Multi-player Online Battle Arena (MOBA)  game \emph{DotA 2} (``Defense of the Ancients''), published by Valve, with around 12.4 million unique players per month (\url{http://blog.dota2.com/}). Thus the contribution presented here is: 

\begin{itemize}
\item We present methods and results for win prediction in professional games using extremely high skill public (casual) game data to supplement the pro-level training data and ensure that the training data covers the data space sufficiently to generate reliable prediction models.
\item We thoroughly evaluate two common prediction algorithms and their  configurations to identify the best performing algorithm and configuration on various aspects of MOBA data. This not only indicates which algorithm and configuration to use and when, but also, how much optimization is required for highest prediction accuracy. 
\end{itemize}

\section{Related Work}\label{sec:Work}
When using data to predict the winner of \emph{DotA 2} matches, the data are sets of instances (vectors) of features representing the game. Features used include pre-game features such as the heroes picked; in-game features such as time-series vectors of game statistics, graphs of hero positions or combat graphs; and post-game features such as hero win rates, player skill scores and player win rates. The machine learning algorithms learn the mapping of these input vectors to output labels (winning team) and then predict the winning team for new data vectors using the learned model. 

There is a wide variety of machine learning algorithms for supervised win prediction in \emph{DotA 2}. The difference lies in how the algorithms build their models and how those models function internally. Much previous work used logistic regression (LR) \cite{eggert2015classification,kinkade2015dota,pobiedina2013ranking,schubert2016esports,song2015predicting,2017arXiv170103162Y}. \citeauthor{Kalyanaraman2015ToWO} \shortcite{Kalyanaraman2015ToWO} compared LR and Random Forests (RF) and found a tendency for RFs to over-fit the training data so they enhanced LR with genetic algorithms. In contrast, \citeauthor{johansson2015result} \shortcite{johansson2015result} found RF had the highest prediction accuracy and that kNN (along with support vector machines) were unsuitable due to the excessive training time (over 12 hours on 15,146 files). \citeauthor{rioult2014mining} \shortcite{rioult2014mining}; \citeauthor{yang2014identifying} \shortcite{yang2014identifying} used decision trees which have the twin advantages of being simple and allowing rules to be extracted while \citeauthor{Semenov2017} \shortcite{Semenov2017} used both Factorization Machines and extreme gradient boosting (Random Forests with meta-learning).


Analyzing the previous academic research on \emph{DotA 2} win prediction, a number of limitations can be identified: \textbf{1) Professional games:} Previous work has not analyzed professional games apart from a small analysis of in-game combat by \cite{yang2014identifying}. Pro game prediction has the highest interest in the esports community.
\textbf{2) Skill rank comparison:} Previous work does not analyze both pro games and non-pro games, or compare pro games to extremely high skill non-pro matches, despite \citeauthor{drachen2014skillbased} \shortcite{drachen2014skillbased} documenting the differences in spatio-temporal behavior of players across skill levels. DotA 2 matches are played across a diverse range of rankings. 
\textbf{3) Metagame:} Previous work has collected match data over time periods that crossed significant changes to the game when new game patches were released, meaning prediction models risk being affected by these changes. These patches alter the ``meta-game'' (the high level strategies adopted by players and teams beyond the standard rule of the game). 

\section{Dataset and Pre-processing}\label{sec:DataSet}
In a \emph{DotA 2} match, there are 10 human players in two teams called ``Dire'' and ``Radiant'' (5 players per team). It is played on a square map split diagonally into two sides by a river. Each side of the map ``belongs'' to a team. Before the match begins, each player picks a unique hero (game character) from 113 possible heroes currently. Each hero has different characteristics. Thus, the combination of heroes on each team can significantly affect which team wins or loses. The more advanced players consider their hero combinations very carefully. Once the match commences, the heroes play different roles where they aim to generate resources to progress through hero levels and become more powerful via fights against the rival team. A team wins when they destroy the other team's nexus (base of structures). Winning a game requires coordination within the team and the ability to react to the opposition's tactics and behavior.

Our \emph{DotA 2} data set has 1,933 replays downloaded 
from Valve's website. Replays consist of low-level game events that occurred when the match was played and are used by DotA 2 engines to recreate entire matches.  We used Clarity, a Java open-source replay parser available at: \url{https://github.com/skadistats/clarity}, to convert the replay file into CSV files of data vectors.  Our data contains 270 pro matches (13.97\% of total) and 1,663 public matches with extremely high skill score ($>$6000 which represents the 99.81 percentile \url{https://dota.rgp.io/mmr/}), played between 27th March 2017 and 4th May 2017.  

A key feature of these data is the mix of professional games and extremely high skill non-pro games. Our aim is to accurately predict professional matches and we need to establish whether high skill public matches can be used as a proxy for professional training data. There are only a limited number of professional matches 
which limits the training data possible, particularly as the mechanics and ``meta'' of the game change significantly when new patches are released. A new patch may mean that previous data is 
potentially irrelevant as the heroes, mechanics and meta have changed and the previous data has to be discarded from the learned model. During our data collection period there were no changes to the core mechanics of the game, such as major patches, which makes this dataset especially appropriate for algorithm development and testing. 

We prepare two datasets; one using pre-match features and one using in-game features. Hero vectors (pre-match features) are the most commonly used data features in the literature. Another popular data feature is time-series vectors of various in-game metrics \cite{Semenov2017}. We split each dataset into training data and testing data. To allow us to evaluate predicting winners from a mix of data versus predicting winners from professional only game data, we use two data splits: all data and tournament data. When analyzing all data, we split the data 66\% for training and 34\% for testing with the data sorted in chronological order. This ensures we never use future data to predict past data which could not happen in reality. To predict tournament data, we use a training data set of all data minus the matches in the 2017 Kiev Major DotA2 tournament (\url{http://dota2.gamepedia.com/Kiev_Major_2017}) and a test set of the 113 matches in the Kiev Major (24-30 Apr. 2017). 

\subsection{Pre-Match Data}\label{sec:heroData}
Our pre-match data comprises 113-dimensional tri-state hero vectors where $x_i$ is 1 if hero $i$ was in Radiant; $x_i$ is -1, if hero $i$ was in Dire and $x_i$ is 0 otherwise. This will allow analysis of the heroes selected and also hero combinations and dependencies where applicable. We refer to the mixed professional and non-professional dataset as \textbf{Mixed-Hero} and the Kiev Major tournament dataset as \textbf{Pro-Hero}.

\subsection{In-Game Data}\label{sec:timeSeriesData}
Our second dataset pair comprises in-game (time-series) data slices from a sliding window of 5-minute intervals. For this evaluation, we use one 5-minute sliding window at the 20 minute (halfway) game time as the average DotA2 game lasts approx. 40 minutes (\url{https://dota.rgp.io/}). We refer to the mixed professional and non-professional dataset as \textbf{Mixed-InGame} and the Kiev Major tournament dataset as \textbf{Pro-InGame}. The machine learning algorithms learn 5-minute sliding window time-series data (convoluted in the time domain) from the 20-minute mark of all matches longer than 20 minutes.
In a 5-minute sliding window, there are 30 individual features each convoluted in the time domain plus the 5 time-stamps and the class label for the vector ``DireWin'' or ``RadiantWin''. We use the following in-game metrics to generate the features of vector $X_{rt}$ to represent the current game state for replay $r$ at time $t$. Each game metric is recalculated for each time stamp $t$. For each metric, we calculate the value for team Dire $D$, the value for team Radiant $R$, the difference between Radiant and Dire $R-D$ and the change (gradient) since the last timestamp for Dire $dD$ and Radiant $dR$ respectively. Table \ref{tab:metrics} lists the metrics. 
\begin{table*}
\centering
\caption{\label{tab:metrics}Details of the in-game metrics used to produce the vectors to train into the machine learning predictors. All vectors are team-based (sum of individual scores) and there is one vector for each timestamp.}
\begin{tabular}{l|p{13.5cm}}
Feature/Metric & Description for each player (summed to give team score) \\\hline
Team Damage Dealt & This represents the amount of damage each player dealt to enemy entities since the game began. \\
Team Kills & The number of enemy heroes killed since the game began.\\
Team Last Hits & The total last hits (who hit last when an enemy entity died). \\
Team Net Worth & Sum of gold in the bank, the value of a player's items in the courier and of those in their inventory.\\
Team Tower Damage & Damage dealt to enemy towers since the game began,\\
Team XP Gained & XP is earned by being within a specific radius of a dying enemy unit.\\
\end{tabular}
\end{table*}

We train a separate winner predictor for each minute through the game. Hence, the learned model $M_t$ at time $t$, is trained with a vector $Xr_t$ representing the game state for replay file $r$ at time $t$ where: $Xr_t = xi_{t - 4}, xi_{t - 3}, ..., xi_{t}$  for all features $i$, and there is one model $M_t$ for each minute interval between 4 and $n$ where $n$ is the maximum game length in minutes. Here we use the 5-minute sliding window for the 20-minute mark which contains \{$xi_{16}$, $xi_{17}$, $xi_{18}$, $xi_{19}$, $xi_{20}$\} for all features $i$.

\section{Evaluation}\label{sec:Eval}
The purpose of this evaluation is to predict \textbf{professional} data using \textbf{mixed} data comprising both professional data and extremely high-skill non-professional data. This will establish whether the mixed data can be used as proxy data for professional data in prediction model building as sufficient professional data is not available for accurate model building.

As noted in section \ref{sec:Work}, Logistic Regression (LR) and Random Forests (RF) \cite{breiman2001random} are popular algorithms for \emph{DotA 2} win prediction. We use them both to analyze our hypothesis that mixed data can be used to accurately predict the winners of professional games. The extremely high level public games in the mixed data act as a proxy for professional games. There are only a limited number of professional games which limits their use for training as the data will not adequately cover the data space for model building.

LR builds a linear model for classification problems. LR estimates the probabilities for each class (DireWin or RadiantWin) using a logistic function of the data features (known as explanatory variables). LR does not consider combinations and dependencies of features. This is particularly germane for heroes where the hero combination is of paramount importance in \emph{DotA2}. LR estimates the importance of individual heroes with respect to the result of the match. 

In contrast to this, RFs are compound decision trees. Each tree in the forest learns a different version of the dataset; equal in size to the training set. This versioned dataset is generated from the original training data using random sampling with replacement (bootstrapping). The versioned dataset will therefore contain some duplicates. RF builds the set of trees by randomly choosing a set of features and then finding the feature within this subset that splits the set of classes optimally. 
To allow the RF to predict, it uses majority voting on the prediction of all trees in the forest, known as bagging. Unlike LR, RFs do consider combinations of heroes as they are essentially rule-based algorithms where the rules are determined by the tree branches.

\subsection{Predicting using Pre-Match Data}
Firstly, we compare prediction accuracy for hero features using the two data sets: \textbf{Mixed-Hero} and \textbf{Pro-Hero}. We trained both datasets (described in section \ref{sec:heroData}) into a LR algorithm and both datasets into a RF algorithm. The accuracy shows if professional data is different to mixed data and which prediction algorithm is most suitable for each. To determine the suitability and performance of the two algorithms, we performed analyses using the Weka data mining environment v3.8 \cite{witten00} and the two predictors: ``Logistic'' and ``RandomForest''. We varied the parameters of both predictors to analyze the accuracies for win prediction across a number of configurations. Parameters for LR and RF are analyzed by comparing the results on the training data set and then used to predict the test data. In all evaluations, we ensured that we compared an equivalent number of algorithm, parameter and feature selections at all stages to ensure fairness. For LR, we varied the ridge in the log-likelihood and for RF we varied the number of trees (iterations in WEKA).  Otherwise, all algorithm parameters were left at their defaults. Once we identified the best configuration for each predictor, we then compared the two algorithms' optimal configurations against each other. 

The parameters also include feature subset selections. \citeauthor{eggert2015classification} \shortcite{eggert2015classification} used the WEKA framework to evaluate three feature selectors. The results indicated that WrapperSubsetEval, a wrapper with best-first search, produced the highest accuracy. We also analyzed this ``wrapper''\cite{kohavi1997wrappers} subset selector which uses the algorithm itself to select the best combination of features. We compared its results to  CfsSubsetEval  \cite{hall1999correlation}, a correlation-based ``filter'' \cite{kohavi1997wrappers} method which greedily selects the optimal feature subset independently of the algorithm. It favors features that are highly correlated to the class but uncorrelated to each other to minimize feature redundancy. CfsSubsetEval is based on information theory and we have demonstrated high accuracy on various datasets in the past \cite{hodge2012binary}.

For each algorithm, we evaluated a range of parameter settings to compare multiple configurations.  Table \ref{tab:HeroResults} shows the accuracy for these configurations. For the hero data, WrapperSubsetEval using BestFirstSearch \cite{witten00} coupled with LR and RF achieved higher accuracy than when LR and RF are run using the features selected by CfsSubsetEval with BestFirstSearch.

\begin{table*}
\centering
\caption{\label{tab:HeroResults}Prediction accuracy of the various configurations of algorithms on the ``mixed'' and the ``professional only'' hero vector data. The highest accuracy is shown in bold for each dataset. The table compares the results for LR and RF with all features, and features selected by a wrapper feature selector.}
\begin{tabular}{l|r|r||r|r}
 & \multicolumn{2}{|c||}{Mixed-Hero} & \multicolumn{2}{|c}{Pro-Hero}\\\hline
Predictor & All & Wrapper Select & All & Wrapper Select  \\\hline
LR & 54.6423 & \textbf{58.7519} &  47.7876 & 50.4425 \\
RF & 53.1202 & 58.2953 &  50.4425 & \textbf{55.7522} \\
\end{tabular}
\end{table*}

For \textbf{Mixed-Hero}, the highest prediction accuracy is 58.75\%, using a LR algorithm with features selected by the Wrapper feature selector. For \textbf{Pro-Hero}, the highest accuracy is 55.75\% using a RF predictor with features selected by the Wrapper feature selector. All perform significantly better than random guess (50\% accuracy).

\subsection{Predicting using In-Game Data}
To allow us to compare prediction accuracy for in-game data, we use the 20-minute data described in section \ref{sec:timeSeriesData} for both \textbf{Mixed-InGame} and \textbf{Pro-InGame}. As with the pre-match data, we trained both sliding window training datasets into a LR algorithm and into a RF predictor. Again, we varied the algorithm parameters to analyze the prediction accuracies of a number of configurations on the test data. For LR, we varied the ridge in the log-likelihood and for RF we varied the number of trees (iterations in WEKA). Furthermore, we compared the algorithm configurations' accuracies using the WEKA framework feature selectors CfsSubsetEval and WrapperSubsetEval with BestFirstSearch.

The accuracies will indicate whether \textbf{Pro-InGame} data is different from \textbf{Mixed-InGame} data with respect to win prediction and which algorithm is most suitable for both data sets. Table \ref{tab:TSResults} shows the accuracy for the various configurations. For the in-game data, LR and RF using CfsSubsetEval feature selection produced higher  accuracy than WrapperSubsetEval feature selection with LR and RF. Conversely, WrapperSubsetEval produced higher accuracy than CfsSubsetEval on the hero data.

\begin{table*}
\centering
\caption{\label{tab:TSResults}Prediction accuracy of the various configurations of algorithms on the ``mixed'' and the ``professional'' in-game data.  The highest \%ge is shown in bold for each dataset. The table compares the results for LR and RF with a single time-series feature, all features and features selected by the CFS feature selector.}
\begin{tabular}{l|r|r|r||r|r|r}
 & \multicolumn{3}{|c||}{Mixed-InGame} & \multicolumn{3}{|c}{Pro-InGame}\\\hline
Predictor & 1-Attr & All & CFS Select & 1-Attr & All & CFS Select  \\\hline
LR & 74.1433 ($Kills_{R-D}$) & 73.3645 & 74.9221 & \textbf{75.2212} ($Kills_{R-D}$) & 70.7965 & 71.6814 \\
RF & 67.757 ($Kills_{R-D}$) & 73.053 & \textbf{76.1682} & 61.0619 ($Kills_{R-D}$) & 66.3717 & 68.1416 \\
\end{tabular}
\end{table*}

For \textbf{Mixed-InGame} data, the highest accuracy is 76.17\%, using a RF algorithm with CFS feature selection. For \textbf{Pro-InGame} data, the highest accuracy is 75.22\% using LR with a single time-series feature.  All perform significantly better than random guess (50\% accuracy). It is difficult to exactly compare with the results in the literature as they all used different skill-level data sets but results are comparable.

\section{Discussion and Analysis}\label{sec:DiscussAnalysis}
Predicting the likely winners of esports games as they progress will provide a simple statistic for the audience making games more accessible but the statistic needs to be accurate to be believable. 
There is insufficient pro data available to use for training as pro matches are infrequent and meta-game changes can make data obsolete. There is only enough pro data for testing. 
Hence, our research question is: ``can mixed training data be used as a proxy for pro training data?''. The results suggest slightly lower accuracy for win prediction in pro test data compared to mixed test data. However, with careful parameter optimization, the results are only slightly worse particularly for the in-game data.

It is clear that the algorithm has to be chosen carefully and that the parameters have to be evaluated thoroughly to identify the optimal configuration. \citeauthor{Semenov2017} \shortcite{Semenov2017} posited that the accuracy of win prediction varies across skill levels and that higher skill games are harder to predict. They did not analyze professional games and we would expect these to be even harder to predict. 
Valve have developed DotA2 to include uncertainty and to make the game less predictable so this will inevitably reduce prediction accuracy. 
\begin{figure}
\centering
\includegraphics[width=7cm]{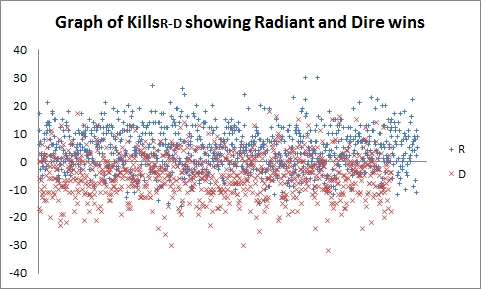}
\caption{Chart of the difference between Radiant kills and Dire kills in the \textbf{Mixed-InGame} data. Above the x-axis is more Radiant kills and below is more Dire kills.} 
\label{fig:allKills}
\end{figure}

One finding of the evaluations summarized in tables \ref{tab:HeroResults} and \ref{tab:TSResults} is that the optimum algorithm varies. For \textbf{Mixed-Hero} data, LR is 0.5\% better but for \textbf{Pro-Hero} data then RF is 5\% better. Conversely, for time-series data, RF is 2.2\% better for \textbf{Mixed-InGame} data yet LR is 5\% better for \textbf{Pro-InGame} data. Least surprising, this suggests that hero data has different characteristics and needs to be treated differently from in-game data supporting other researchers' findings \cite{Semenov2017,2017arXiv170103162Y}. More surprisingly, this suggests that predicting pro data needs to be treated differently compared to predicting the mixed data, even requiring different prediction algorithms. This may indicate that pro players generate different data than non-pro and may also indicate over-fitting by the algorithms.

The best feature selector is different too. For hero data, the wrapper selector outperforms the filter selector. Selecting heroes is more accurate when heroes are chosen with the algorithm (wrapper) rather than independently (filter) as picking is optimized for the algorithm. In contrast, for the in-game data, the filter selector outperforms the wrapper selector. We would expect this as there are feature correlations, (e.g., XP gained and kills in table \ref{tab:metrics} are correlated) and CfsSubsetEval favors feature subsets that are highly correlated to the class but uncorrelated to each other to minimize redundancy. Wrapper methods do not consider correlations.
\begin{figure}
\centering
\includegraphics[width=7cm]{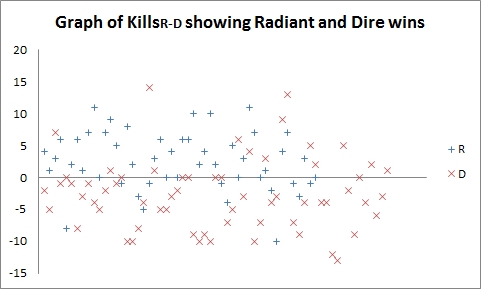}
\caption{Chart showing the difference between the Radiant kills and the Dire kills in the \textbf{Pro-InGame} data. Above the x-axis is more Radiant kills and below is more Dire kills.}
\label{fig:kievKills}
\end{figure}

The best performance is usually achieved by selecting feature subsets but for \textbf{Pro-InGame} data, a single time-series feature, $Kills_{R-D}$, performs best. Figures \ref{fig:allKills} and \ref{fig:kievKills} plot $Kills_{R-D}$ at the 20-minute mark for \textbf{Mixed-InGame} and \textbf{Pro-InGame} respectively. The winning team is shown as blue `+'=Radiant or red `x'=Dire. We would expect the  `+' to be above the x-axis as $R-D$ is positive above the x-axis indicating Radiant had more kills and the `+' indicates Radiant wins. Where the `x' is below the x-axis, Dire had more kills so these should be Dire wins (`x'). Statistically analyzing the \%ge of Radiant wins above the x-axis and Dire wins below , \textbf{Pro-InGame} is 63\% Radiant above and 66\% Dire below. In comparison, \textbf{Mixed-InGame} is 69\% Radiant above and 70\% Dire below. Hence, we would not expect $Kills_{R-D}$ for \textbf{Pro-InGame} to have higher accuracy. This suggests that there is a fundamental difference between \textbf{Mixed-InGame} and \textbf{Pro-InGame} data and the required predictor needs to be evaluated thoroughly.

The accuracy for the hero data is much lower ($<$58.8\%) than the in-game data (up to 76.2\%). Our hero data only consider the sets of heroes selected but not which players were playing those heroes. \citeauthor{2017arXiv170103162Y} \shortcite{2017arXiv170103162Y} posited that which player is playing each hero is very important. In all skill levels including professional, players have preferred heroes and hero types which they play much more competently. Factoring this into the model will improve accuracy. 

\begin{figure}
\centering
\includegraphics[width=7cm]{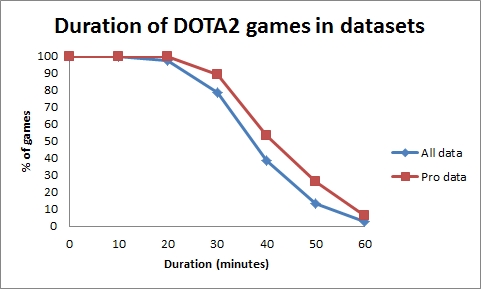}
\caption{Chart showing the duration in minutes of the games as a \%ge of total games in the two datasets analyzed.}
\label{fig:duration}
\end{figure}

In-game metrics represent the game state; i.e., how the game is progressing. They inevitably boost accuracy for win prediction as they effectively signify who is currently leading at each timestamp. We analyze prediction at 20-minutes which is half-way on average. The further the game progresses, the more accurate the in-game predictor will become. \citeauthor{2017arXiv170103162Y} \shortcite{2017arXiv170103162Y} suggested that actions in the later stages have more influence on who wins a game. However, they and \citeauthor{johansson2015result} \shortcite{johansson2015result} both identified that the longer a match lasts then the lower the prediction accuracy throughout as longer matches are more unpredictable. Figure \ref{fig:duration} illustrates the length of \emph{DotA 2} matches in our dataset and clearly shows that professional games last longer and, hence, will be more unpredictable. Considering all games, 97.6\% (1887) of the 1933 games last 20 minutes or longer while 100\% (113) of the professional games last 20 minutes or longer so we would expect these professional games to be more difficult to predict at 20 minutes. 

Analyzing hero vectors further, the accuracy difference between \textbf{Pro-Hero} and \textbf{Mixed-Hero} data is much larger than  the difference between \textbf{Pro-InGame} and \textbf{Mixed-InGame}. Professional players invest time and thought into the hero pick and have developed a broad range of strategies specifically to counter the opposition's selection. Hence, using non-professional data to predict professional hero picks has lower accuracy as the hero picks by professional do not follow standard patterns but are derived from background research and team strategizing. In contrast, the time-series data represent the current game state and indicate which team is ahead at each timestamp. This indicator works for both non-professional and professional data much more closely.

\section{Conclusion and Future Work}\label{sec:Conclusion}
Watching professional esports is becoming a popular social activity \cite{superdata:17}. However, professional games are fast-paced and not easy to understand so viewers need assistance to comprehend the on-screen action. Even casual players need professional games explaining \cite{schubert2016esports}. To make esports more understandable and to broaden its appeal, broadcasters can provide in-game statistics to improve the spectator experience. Predicting the likely winners of games as they progress provides a simple, easily understood in-game statistic for the audience. Existing research into win prediction for MOBAs has been applied across a range of skill levels, but there is no prior work on predicting professional games in \emph{Dota 2}. The ability to accurately predict professional games has the highest potential impact due to the number of viewers and high financial stakes \cite{eggert2015classification,Semenov2017}.

The purpose of this research was to explain professional matches to the audience as the matches progress by accurately predicting the winner using machine learning. As there are insufficient professional matches for training our models, we aimed to supplement professional data with extremely high skill non-professional data to ensure sufficient data for training. Our research question was thus: can we use mixed professional and non-professional training data to predict the winners of professional matches? 

For in-game data, the accuracy when predicting the winner of professional matches from models generated with mixed training data is only slightly 
lower than when predicting mixed data from the same models. The hero data does not perform well enough for further consideration. We can use the mixed in-game data as a basis for a framework to predict the winners of professional matches. This is important as there are insufficient professional games played to generate enough data for training models, particularly as the game and meta-game are constantly evolving.  Our evaluation clearly demonstrates that evaluating multiple machine learning algorithms coupled with algorithm optimization such as feature selection and parameter optimization is vital and a broad range of configurations need to be evaluated to ensure maximum accuracy. In fact, the professional data actually require different algorithms compared to the non-professional data for highest accuracy which may indicate that pro data is different from mixed data and that the models are over-fitting. 

Our approach significantly increases the data availability, enabling for the first time the ability to predict professional games. It marks a transition for professional esports analytics from descriptive to predictive statistics. This helps explain esports for the audience making previously complex games more accessible. 

In future work, we will build on this baseline by analyzing more matches, more features (such as those listed in section \ref{sec:Work}) then explore applying our win prediction methods to digital games more broadly to maximize player and audience engagement. In time, as similar high-frequency detailed datasets become available from the Internet of Things \cite{sun2016internet}, we can begin to explore the application of live prediction to human behavioral data in the real world.

\bibliography{LitRev}
\bibliographystyle{arxiv}
\end{document}